\documentclass[11pt]{article}

\usepackage[final]{acl}

\usepackage[most]{tcolorbox} 
\usepackage{booktabs}
\usepackage{multirow}
\usepackage{graphicx}
\usepackage[section]{placeins}

\usepackage{times}
\usepackage{latexsym}
\usepackage{amsmath,amssymb,amsfonts}

\usepackage[T1]{fontenc}
\usepackage[utf8]{inputenc}

\usepackage{makecell,tabularx,array}
\usepackage{enumitem}
\usepackage{microtype}
\usepackage{inconsolata}

\title{
Do Implicit Personalization and Explicit Styles Conflict?\\
PsPLUG: A Lightweight Plug-in for Balancing Personalization and Style in Customized LLMs
}

\author{
\textbf{Yutong Song}$^{\clubsuit *}$,
\textbf{Jiang Wu}$^{\spadesuit *}$,
\textbf{Shaofan Yuan}$^{\diamondsuit}$,
\textbf{Chengze Shen}$^{\diamondsuit}$, \\
\textbf{Jian Wang}$^{\diamondsuit}$,
\textbf{Yu Wang}$^{\diamondsuit}$,
\textbf{Nikil Dutt}$^{\clubsuit}$,
\textbf{Amir M. Rahmani}$^{\clubsuit}$
\\[0.3em]
$\clubsuit$ University of California, Irvine
\quad
$\spadesuit$ Independent Researcher
\quad
$\diamondsuit$ TikTok
}

\begin{document}

\maketitle

\begingroup
\renewcommand{\thefootnote}{*}
\footnotetext{Equal contribution.}
\endgroup

\begingroup
\renewcommand{\thefootnote}{}
\footnotetext{The code is available at \url{https://github.com/RainieLLM/PsPLUG}.}
\footnotetext{This paper has been accepted by EMNLP 2026 (main).}
\addtocounter{footnote}{-1}
\endgroup

\begin{abstract}

Personalized large language models are often expected to follow explicit style instructions, yet we find that such instructions can undermine the user-specific characteristics that personalization methods aim to preserve. We call this failure mode personalization collapse: explicit style control can conflict with implicit user preferences. To address this challenge, we propose PsPLUG, a lightweight plug-in that learns a user-specific residual after accounting for the requested style. PsPLUG also allows us to tune personalization strength at inference time. Our experiments show that explicit style instructions can diminish personalization in existing methods, whereas PsPLUG better preserves user preferences while providing precise control over the balance between personalization and style adherence.





\end{abstract}

\section{Introduction}

Large language models (LLMs) are increasingly deployed in interactive settings where users expect systems not only to produce factually correct content, but also to align with their individual linguistic habits, preferences, and communicative styles~\cite{tan2023usermodeling,zhang2024personalization}. This has sparked rapid progress in personalized generation, including (i) \emph{retrieval-based} methods that fetch user histories into the context window~\cite{lamp,longlamp,pearl,stepback}, (ii) \emph{per-user fine-tuning} approaches such as user-specific LoRA or adapters~\cite{hu2022lora,houlsby2019adapter,oppu,perpcs}, and (iii) lightweight plug-in mechanisms that inject user embeddings or soft prompts~\cite{liu2024personaplug,li2021prefixtuning,liu2022ptuning}. Together, these techniques have demonstrated that LLMs can adapt to a user given enough data or context.~\cite{chen2024personapersonalizationsurveyroleplaying,pad2025,zhang2025personalized}
\begin{figure}[t!]
    \centering 
    \includegraphics[width=1\linewidth]{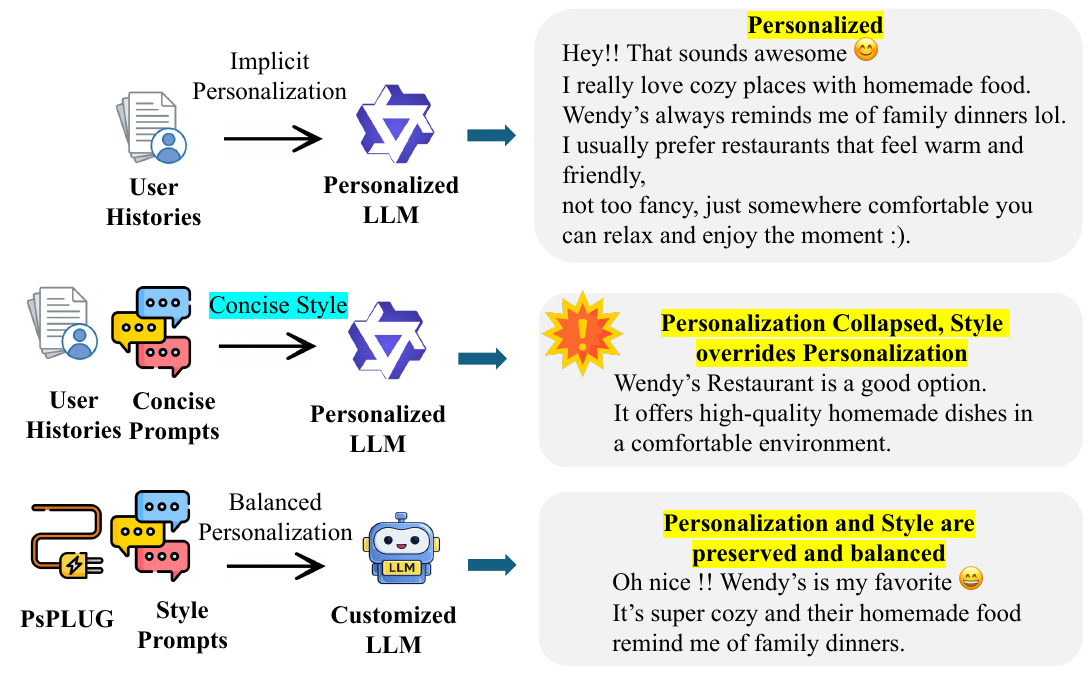} 
    \caption{The style-constrained personalization challenge. While standard models capture implicit user personalization, existing methods suffer from severe personalization collapse under explicit style instructions.}
    \label{fig:psplug_architecture}
\end{figure}
Despite this progress, a critical vulnerability in current personalization pipelines remains largely overlooked. Existing methods typically inject user-related signals without theoretically clarifying what constitutes the \emph{core} personalized signal, or how it relates to the neutral behavior of the base model under the same inputs~\cite{pref,drift}. Concurrently, modern NLP applications increasingly operate under \emph{explicit system instructions}, such as strict stylistic or tonal guidance (e.g., ``respond formally'', ``use a concise tone'')~\cite{zhang2023instruction,liang2024controllable,shanahan2023roleplaylargelanguagemodels}. We empirically observe that when such explicit constraints are introduced, existing personalization methods suffer from severe persona degradation. Strong system instructions tend to dominate the generation space, effectively overriding and collapsing diverse dimensions of user-specific traits. This reveals a fundamental challenge: \emph{style-constrained personalization}, where a model must reliably balance explicit task directives with implicit user priors.

To address this, we introduce a novel theoretical perspective: modeling personalization as a \emph{distributional residual}. Rather than learning absolute output likelihoods, we view the persona as the distinct deviation between two conditional distributions under the \emph{identical} input: the user’s true linguistic distribution and the neutral distribution of a base LLM. A user-authored response naturally encapsulates personalized lexical and structural preferences, whereas a zero-shot LLM response reflects a generic, population-level prior. Their divergence constitutes the pure personalization residual. 

Building on this residual view, we design a style-aware preference objective~\cite{bradley1952rank,rafailov2023direct}. By contrasting user-authored texts against style-conditioned negatives—responses following explicit instructions but lacking user nuances—the model isolates latent persona signals. Unlike standard preference optimization that merely anchors to generic base distributions (e.g., LongPO~\cite{chen2025longpolongcontextselfevolution}), our residual reward mathematically disentangles these competing signals to preserve fine-grained persona fidelity under strict stylistic constraints.

We instantiate this methodology as \textbf{PsPLUG}, a lightweight soft-prompt module that prepends learned prefix embeddings to a frozen LLM backbone. To achieve a controllable equilibrium between dual constraints, PsPLUG incorporates a unified inference-time scaling mechanism governed by a coefficient ($\alpha$). This grants fine-grained, dynamic adjustment over the trade-off between instruction adherence and personalization strength, enabling scalable adaptation without the prohibitive costs of per-user fine-tuning.

Our work makes three main contributions:
\begin{itemize}
\item We propose PsPLUG, a novel framework that effectively balances explicit system instructions with implicit user personalization, addressing the critical issue that existing methods often suffer from severe personalization degradation during text generation, as strong system instructions tend to override and collapse diverse dimensions of user-specific traits. 

\item We introduce a novel perspective that models personalization as a distributional residual. Building on this view, PsPLUG uses a style-conditioned preference objective to separate user-specific persona signals from dominant instruction and style effects, enabling parameter-efficient personalization without per-user fine-tuning.

\item We develop a unified inference-time control mechanism with a scaling coefficient ($\alpha$), which allows fine-grained adjustment of the trade-off between instructions and personalization strength. Experiments on the LaMP benchmark show that PsPLUG consistently outperforms state-of-the-art baselines in preserving persona alignment under strong instructions.
\end{itemize}

\section{Method}
\subsection{Task Formulation}

Let $x$ denote a task input (e.g., a news headline prompt), $u$ index a user, and $y^{u}$ be the user-authored response.
Let $s$ denote an optional style instruction, and define the full task input as $x_s \triangleq (x,s)$.
We assume access to a base LLM $\pi_{\mathrm{ref}}(y \mid x_s)$ that represents a non-personalized, population-level distribution.
Our goal is to construct a personalized policy $\pi_{\phi}(y \mid x_s, u)$ that (i) reproduces user-specific behavior and (ii) can be balanced with explicit task-level style instruction $s$.
\begin{figure*}[t]
  \centering
  \includegraphics[width=1\linewidth]{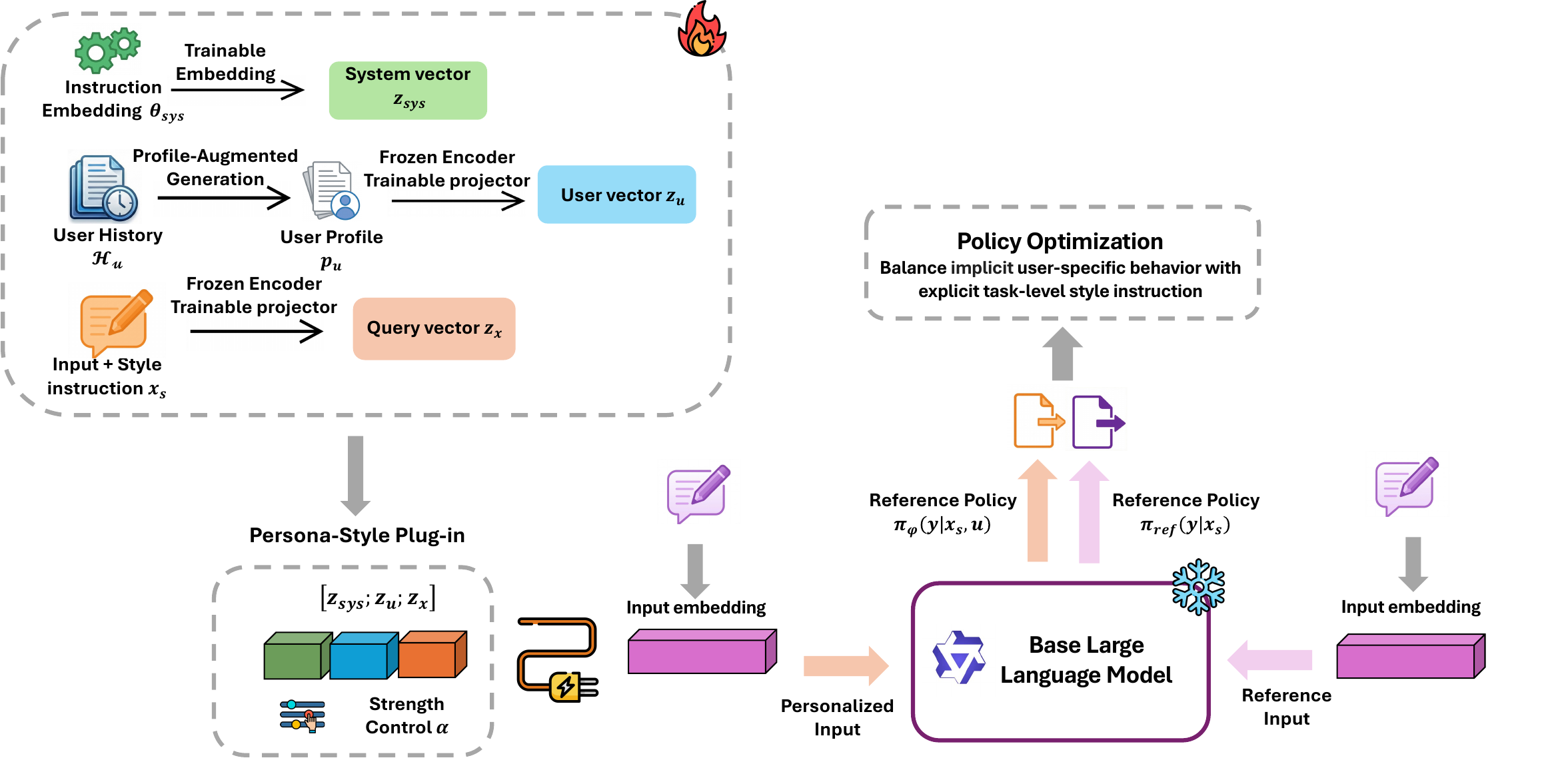}
  \caption{PsPLUG is a lightweight plug-in framework that injects a learnable user-specific prefix into a frozen LLM to enable controllable personalization. The fire logo indicates the trainable model and the snow logo indicates the frozen model.}
  \label{fig:main}
\end{figure*}
In our experiments, we consider a fixed set of four predefined style instructions summarized in the Table \ref{tab:style-prompts}.

\begin{table}[t]
\centering
\footnotesize
\caption{Predefined style instructions used in PsPLUG training and evaluation.}
\renewcommand{\arraystretch}{0.95}
\begin{tabularx}{\linewidth}{@{} l X @{}} 
\toprule
\textbf{Style Name} & \textbf{Style Instruction} \\
\midrule
Warm &
Please write in a warm, humorous style that uses gentle jokes and soft, uplifting comedy. \\
\addlinespace[3pt] 
Critical &
Please write in a sharply critical way, directly pointing out flaws or problems and avoiding overly balanced phrasing. \\
\addlinespace[3pt]
Concise &
Please write in a concise and formal way, using precise language and avoiding unnecessary elaboration. \\
\addlinespace[3pt]
Elaborative &
Please write in a reflective and elaborative way, carefully explaining reasoning with detailed examples and considering multiple perspectives.\\
\bottomrule
\end{tabularx}
\label{tab:style-prompts}
\end{table}

\subsection{PsPLUG Framework}
We instantiate the residual view with a lightweight plug-in that injects a compact continuous prefix into a frozen LLM.
PsPLUG is trained once and attached to task prompts at inference time.
Crucially, because the backbone remains frozen and user signals are decoupled into a modular prefix, our approach enables  rapid user context switching without the need to reload model parameters.
Furthermore, scaling this prefix provides a continuous \emph{strength control mechanism}, yielding a tunable trade-off between personalization and style instructions.

We denote by $\phi$ all trainable plug-in parameters, including the shared instruction embedding $\theta_{\mathrm{sys}}$ and the parameters of $f_{\phi}^{\mathrm{pag}}$ and $f_{\phi}^{\mathrm{qry}}$. The backbone $\pi_{\mathrm{ref}}$ (including its input embedding layer $\mathrm{Emb}_{\mathrm{ref}}$) remains frozen.

Formally, we inject a compact prefix $z_{u,x_s}$ at the input embedding layer. This prefix is the concatenation of three distinct vectors:
(i) a trainable system instruction vector $z_{\mathrm{sys}}$;
(ii) a user vector $z_u$ derived from the user's history;
(iii) an input vector $z_x$ derived from the current prompt.
The personalized policy is defined as:
\begin{equation}
\begin{aligned}
z_{u,x_s} &\triangleq [z_{\mathrm{sys}};\, z_u;\, z_x],\\
\pi_{\phi}(y\mid x_s,u)
&= \pi_{\mathrm{ref}}\!\big(y\mid [z_{u,x_s};\, \mathrm{Emb}_{\mathrm{ref}}(x_s)]\big).
\end{aligned}
\label{eq:soft-prompt}
\end{equation}

\paragraph{System Instruction Embedding.}
Besides user- and input-dependent components, PsPLUG includes a shared system instruction embedding that is global across users.
This design is inspired by recent studies on instruction tuning~\cite{su2023instruction,zhang2023instruction}, which demonstrate that modeling instructions as continuous embeddings can improve an LLM’s ability to interpret and follow task-level guidance.
We parameterize this component as a trainable embedding $\theta_{\mathrm{sys}}\in\mathbb{R}^{d}$ and share it across all users:
\begin{equation}
    z_{\mathrm{sys}} \triangleq \gamma\,\theta_{\mathrm{sys}},
    \label{eq:z-sys}
\end{equation}
where $\gamma$ is a constant embedding-scale factor (we set it to match the typical norm of the backbone input embeddings).
The shared instruction embedding $\theta_{\mathrm{sys}}$ is optimized jointly with the other plug-in parameters in $\phi$.
At inference time, $z_{\mathrm{sys}}$ is prepended to the input embedding sequence, serving as a shared instruction anchor across users.

\paragraph{Personalization Embedding Projector.}
To inject user-specific signals, we convert the user history $\mathcal{H}_u$ into a short fixed Profile-Augmented Generation (PAG) text descriptor $p_u$, and encode it with a frozen sentence encoder $E(\cdot)$:
\begin{equation}
    p_u \triangleq \mathrm{PAG}(\mathcal{H}_u), \qquad
    e_u \triangleq E(p_u).
    \label{eq:pag-embed}
\end{equation}
We cache $e_u$ offline and then map $e_u$ into the LLM hidden space through a trainable multi-layer perceptron (MLP) projector:
\begin{equation}
    z_u \triangleq \gamma\, f_{\phi}^{\mathrm{pag}}(e_u),
    \label{eq:z-user}
\end{equation}
where $\gamma$ is an embedding-scale factor; during training, $f_{\phi}^{\mathrm{pag}}$ is optimized as part of the plug-in parameters $\phi$, while $E(\cdot)$ and the backbone remain frozen.

\paragraph{User Query Encoder.}
User preferences can manifest differently depending on the current request, so PsPLUG further includes an input-aware component.
Given the current input prompt $x_s$, we obtain a fixed input feature by pooling token embeddings from the frozen backbone:
\begin{equation}
    h(x_s) \triangleq \mathrm{Pool}\!\big(\mathrm{Emb}_{\mathrm{ref}}(x_s)\big).
    \label{eq:hx}
\end{equation}
We then map $h(x_s)$ into the hidden space with a trainable MLP encoder:
\begin{equation}
    z_x \triangleq \gamma\, f_{\phi}^{\mathrm{qry}}(h(x_s)).
    \label{eq:z-input}
\end{equation}
Since the backbone is frozen, $h(x_s)$ is non-trainable; we only optimize the plug-in parameters $\phi$ (including $f_{\phi}^{\mathrm{qry}}$, $f_{\phi}^{\mathrm{pag}}$, and $\theta_{\mathrm{sys}}$).

\subsection{PsPLUG Training: Persona--Style Balancing}
\label{sec:style-dpo}

Consider a generic input $x$ augmented with a style instruction $s$
(e.g., ``write in a concise formal style''), represented as
$x_s \triangleq (x,s)$.

This formulation highlights a fundamental tension and raises two questions:
(i) Can the model preserve personalization without violating the style instruction?
(ii) Can we explicitly \emph{control} the intensity of personalization relative to the style?

\paragraph{Construction of Style-Conditioned Pairs.}
To isolate user-specific signals from generic style following, we construct preference pairs under the style-augmented context $x_s$.
We first generate a \emph{style-only} baseline from the model:
\begin{equation}
    y^{s} \sim \pi_{\mathrm{ref}}(\cdot \mid x_s).
\end{equation}
Since $y^{s}$ follows $s$ but contains no user-specific information, we treat the user-authored response $y^{u}$ as the preferred output and form the style-conditioned pair $(y^+,y^-)\triangleq(y^{u},y^{s})$.
This construction ensures that the learning signal focuses on the user residual beyond style compliance.

\paragraph{Residual Based Training Objective.}
PsPLUG therefore models personalization as learning a \emph{residual} between two conditional distributions under the same context $x_s$:
the personalized policy $\pi_\phi(\cdot\mid x_s,u)$ and the style-conditioned reference $\pi_{\mathrm{ref}}(\cdot\mid x_s)$.
Using the preference pair $(y^{+},y^{-})$ constructed above, we optimize $\phi$ via a Bradley--Terry (BT) pairwise loss \cite{bradley1952rank}.
For brevity, we omit $(x_s,u)$ in the notation and define the implicit reward score $r_\phi(y)$ as the log-likelihood ratio:
\begin{equation}
    r_\phi(y)\triangleq \log \pi_\phi(y\mid x_s,u) - \log \pi_{\mathrm{ref}}(y\mid x_s).
\label{eq:reward}
\end{equation}
The loss function is defined as:
\begin{equation}
\begin{aligned}
    \ell_{\mathrm{style}}(\phi; x_s, u, y^{+}, y^{-})
    &= -\log \sigma\!\big(\beta\,\Delta r_\phi\big), \\
    \Delta r_\phi
    &\triangleq r_\phi(y^{+}) - r_\phi(y^{-}).
\end{aligned}
\label{eq:bt-loss}
\end{equation}

where $\sigma(\cdot)$ is the sigmoid function and $\beta>0$ is a temperature hyperparameter.
Here, the reference terms anchor the comparison to the baseline style-following behavior induced by $s$ (captured by $\pi_{\mathrm{ref}}$), while the personalized policy is encouraged to deviate only when it better captures the user's specific preferences ($y^u$) over the generic style ($y^s$).

\paragraph{Inference Time Personalization Strength Control.}
Because PsPLUG is modular, we can explicitly regulate the intensity of the user-specific signal during inference without affecting the system instruction or input understanding.
We introduce a scaling coefficient $\alpha \ge 0$ specifically for the user vector $z_u$.
Let $z_{u,x_s} = [z_{\mathrm{sys}}; z_u; z_x]$ be the learned prefix components. The control mechanism is applied as follows:
\begin{equation}
\begin{aligned}
    z^{(\alpha)}_{u,x_s} &= [z_{\mathrm{sys}};\, \alpha \cdot z_u;\, z_x], \\
    \pi_{\phi}^{(\alpha)}(y\mid x_s,u)
    &= \pi_{\mathrm{ref}}\!\big(y\mid [z^{(\alpha)}_{u,x_s};\, \mathrm{Emb}_{\mathrm{ref}}(x_s)]\big).
\end{aligned}
\label{eq:alpha-control}
\end{equation}
By scaling only the user history projector output $z_u$, $\alpha$ serves as a continuous control parameter for personalization strength:
as $\alpha \to 0$, the user-specific signal is progressively suppressed, whereas $\alpha>1$ amplifies the influence of the user's historical personalization.

\subsection{Special Case: Personalization without Style Instructions}
\label{sec:neutral-dpo}

The style-guided formulation in Section~\ref{sec:style-dpo} admits a natural special case when no explicit style instruction is provided.
Formally, setting $s=\emptyset$ reduces the conditioning context $x_s=(x,s)$ to the raw input $x$,
and the reference policy $\pi_{\mathrm{ref}}(\cdot\mid x_s)$ collapses to a generic baseline as $\pi_{\mathrm{ref}}(\cdot\mid x)$.

\paragraph{Neutral Baseline and Preference Pairs.}
Under this setting, the style-only baseline $y^{s}$ in Section~\ref{sec:style-dpo} reduces to a \emph{neutral baseline}:
\begin{equation}
    y^{0} \sim \pi_{\mathrm{ref}}(\cdot \mid x).
\end{equation}
The preference pair becomes $(y^{+},y^{-})\triangleq(y^{u},y^{0})$.
This construction isolates the user-specific residual relative to generic population behavior.

\paragraph{Residual Objective without Style.}
Substituting $s=\emptyset$ into the implicit reward score in Eq.~\ref{eq:reward}, we obtain:
\begin{equation}
    r_\phi(y) = \log \pi_\phi(y\mid x,u) - \log \pi_{\mathrm{ref}}(y\mid x).
\end{equation}
The corresponding Bradley--Terry loss is:
\begin{equation}
\begin{aligned}
    \ell_{\mathrm{neutral}}&(\phi; x,u,y^{u},y^{0}) \\
    &= - \log \sigma \Big( \beta \big[ r_\phi(y^{u}) - r_\phi(y^{0}) \big] \Big).
\end{aligned}
\label{eq:neutral-bt}
\end{equation}
This special case allows PsPLUG to learn personalization with no style guidance.
Gradients update the plug-in parameters $\phi$, while user-dependent information is injected only via $z_u$, which thus serves as the main carrier of the personalization residual relative to $\pi_{\mathrm{ref}}$.

\section{Experimental Settings}

\subsection{Datasets and Evaluation}
We follow the official LaMP benchmark protocol.
We report F1 and accuracy for LaMP-1, accuracy and F1 for LaMP-2, MAE and RMSE for LaMP-3,
and ROUGE-1 / ROUGE-L / METEOR ~\cite{lin-2004-rouge} for LaMP-4, LaMP-5, and LaMP-7.
In the style text generation experiments, we consider a fixed set of four predefined style instructions,
summarized in Table~\ref{tab:style-prompts}. More task and dataset split details are shown in Appendix~\ref{app:dataset}.

In addition to the task metrics, we report two auxiliary scores for controlled customization:
\emph{personalization-score} measures alignment with the user’s preferences, and \emph{style-score} measures adherence to the given style instruction.
Both are computed using LLM-based judges, with human validation on a subset.
Details of the judging protocol are provided in Appendix~\ref{app:system_prompts}.

\subsection{Implementation Details}
Across all tasks, we use Qwen/Qwen3-8B~\cite{yang2025qwen3technicalreport} as the backbone LLM.
For each user, we construct a PAG profile via greedy decoding using vLLM~\cite{kwon2025vllm}.
The profile text is encoded by a frozen sentence encoder (BGE-base-en-v1.5) using the \texttt{[CLS]} representation with $\ell_2$ normalization, and the resulting embeddings are cached offline.
For evaluation, we employ GPT-5.2 PRO as an automated LLM judge to assess generation quality.
All experiments are conducted on 8 NVIDIA H100 GPUs, and full hyperparameter settings are provided in Appendix~\ref{app:hyperparameters}.

\subsection{Baselines}
We compare PsPLUG with the following baselines (see Appendix~\ref{app:detailed_baselines} for implementation details):

\paragraph{Non-personalized}
The LLM generates outputs conditioned solely on the task input, without accessing user history.

\noindent\textbf{Naive retrieval-based personalization.}
We implement a retrieval-augmented baseline that retrieves the top-$k$ user history items via BM25~\cite{robertson2009bm25} and prepends them to the input as demonstrations.

\noindent\textbf{State-of-the-Art Personalization.}
We include three representative methods: \textbf{PAG}~\cite{pag}, \textbf{OPPU}~\cite{oppu}, and \textbf{PPlug} (Persona-Plug)~\cite{liu2024personaplug}. These methods align the backbone LLM with user interests using learnable soft prompts or plug-in modules. We match the backbone and decoding configurations for a fair comparison.

\footnotetext{\textsuperscript{*}Denotes one-time pre-processing. $|P_u|$: size of user history; $H, L$: backbone hidden size and layers; $r$: LoRA rank; $d_e$: embedding dimension. Note that PsPLUG injects fixed vectors, keeping inference overhead constant and independent of $|P_u|$.}

\section{Results and Analysis}

\paragraph{Research Questions}
In this section, we present comprehensive experiments, aiming to address the following Research Questions (RQs):\\
\noindent\textbf{RQ1:} How does \textsc{PsPLUG} perform on personalization tasks without explicit style instructions compared to existing personalization baselines?

\noindent\textbf{RQ2:} Can \textsc{PsPLUG} maintain effective user-level personalization when explicit style constraints are introduced in text generation tasks? 


\noindent\textbf{RQ3}: Can \textsc{PsPLUG} achieve a controllable trade-off between stylistic adherence and personalized expression?

\noindent\textbf{RQ4}: Is \textsc{PsPLUG} a lightweight and efficient plug-in approach, and can it adapt across base LLMs of different model sizes?

\subsection{Main Results}
To answer \textbf{RQ1}, we compare PsPLUG with other personalized baselines and results are shown in Table \ref{tab: results DPO}. \textbf{Consistently outperforms existing personalization baselines on tasks without explicit style instructions.} While a few tasks (e.g., LaMP-5) remain competitive with PPlug, PsPLUG maintains comparable performance overall.

\begin{table*}[t]
\centering
\caption{Main personalization results on LaMP benchmarks. $\uparrow$ indicates higher is better, and $\downarrow$ indicates lower is better. Best results are in \textbf{bold}, and second-best are \underline{underlined}.}
\label{tab: results DPO}
\small
\setlength{\tabcolsep}{3.5pt}
\renewcommand{\arraystretch}{0.95}
\begin{tabular}{llcccccc}
\toprule
\textbf{TASK/methods} & \textbf{Metric} & \textbf{Non-per.} & \textbf{RAG} & \textbf{PAG} & \textbf{PPlug} & \textbf{OPPU} & \textbf{PsPLUG (Ours)} \\
\midrule

\multirow{2}{*}{\makecell[l]{\textbf{LaMP-1:}\\\textbf{CITATION ID.}}}
& ACC$\uparrow$ & 0.518 & 0.441 & 0.562 & \underline{0.563} & 0.556 & \textbf{0.584} \\
& F1$\uparrow$  & 0.448 & 0.397 & 0.481 & 0.493 & \underline{0.556} & \textbf{0.589} \\
\midrule

\multirow{2}{*}{\makecell[l]{\textbf{LaMP-2M:}\\\textbf{MOVIE TAGGING}}}
& F1$\uparrow$  & 0.254 & 0.283 & 0.311 & 0.307 & \underline{0.314} & \textbf{0.334} \\
& ACC$\uparrow$ & 0.375 & 0.381 & \underline{0.393} & 0.382 & \textbf{0.425} & 0.392 \\
\midrule

\multirow{2}{*}{\makecell[l]{\textbf{LaMP-3:}\\\textbf{PRODUCT RATING}}}
& MAE$\downarrow$  & 0.516 & 0.615 & 0.435 & \underline{0.339} & 0.347 & \textbf{0.332} \\
& RMSE$\downarrow$ & 0.805 & 0.981 & 0.714 & \underline{0.583} & 0.613 & \textbf{0.464} \\
\midrule

\multirow{3}{*}{\makecell[l]{\textbf{LaMP-4:}\\\textbf{NEWS HEADLINE GEN.}}}
& ROUGE-1$\uparrow$ & 0.146 & \underline{0.165} & 0.164 & 0.158 & 0.152 & \textbf{0.167} \\
& ROUGE-L$\uparrow$ & 0.128 & 0.144 & \underline{0.146} & 0.138 & 0.128 &  \textbf{0.148}\\
& METEOR$\uparrow$  & 0.107 & \underline{0.108} &0.098  & 0.092 & 0.079 & \textbf{0.109} \\
\midrule

\multirow{3}{*}{\makecell[l]{\textbf{LaMP-5:}\\\textbf{SCHOLARLY TITLE GEN.}}}
& ROUGE-1$\uparrow$ & 0.426 & 0.459 & 0.415 & \underline{0.462} & 0.426 & \textbf{0.463} \\
& ROUGE-L$\uparrow$ & 0.342 & \underline{0.387} & 0.352 & 0.386 & 0.342 & \textbf{0.391} \\
& METEOR$\uparrow$  & 0.360 & \textbf{0.401} & 0.375 & \underline{0.399} & 0.393 & 0.398 \\
\midrule

\multirow{3}{*}{\makecell[l]{\textbf{LaMP-7:}\\\textbf{TWEET PARAPHRASE}}}
& ROUGE-1$\uparrow$ & 0.497 & 0.500 & \underline{0.507} & 0.502 & 0.498 & \textbf{0.523} \\
& ROUGE-L$\uparrow$ & 0.440 & 0.441 & 0.435 & \underline{0.443} & 0.422 & \textbf{0.457} \\
& METEOR$\uparrow$  & 0.324 & \textbf{0.337} & 0.329 & \underline{0.332} & 0.327 & \textbf{0.337} \\
\bottomrule
\end{tabular}
\end{table*}

To answer \textbf{RQ2}, we compare PsPLUG with other 
personalized baselines under $4$ style prompt and results are shown in Table \ref{tab:style_lamp}. We have findings as follows.

\textbf{PsPLUG maintains strong user-level personalization under explicit style constraints}. It achieves best or second-best performance in over 80\% of style–task–metric settings and outperforming prior baselines by up to 2–4 ROUGE points.

\textbf{Different styles exhibit varying degrees of interference with personalization.} Tone-oriented styles such as warm and critical tend to disrupt personalization baselines more severely, often leading to noticeable performance degradation.
In contrast, \textbf{Under the concise constraint, some baselines even improve score} (e.g., Non-pers. baseline R-1 on LaMP-5 increases from 0.426 to 0.441), suggesting that concise is inherently aligned with task structure and user preferences rather than conflicting with them. This behavior suggests that concise primarily functions as a structural constraint on content length and organization, rather than altering tone or sentiment. 
\begin{figure*}
  \centering
  \includegraphics[width=1\linewidth]{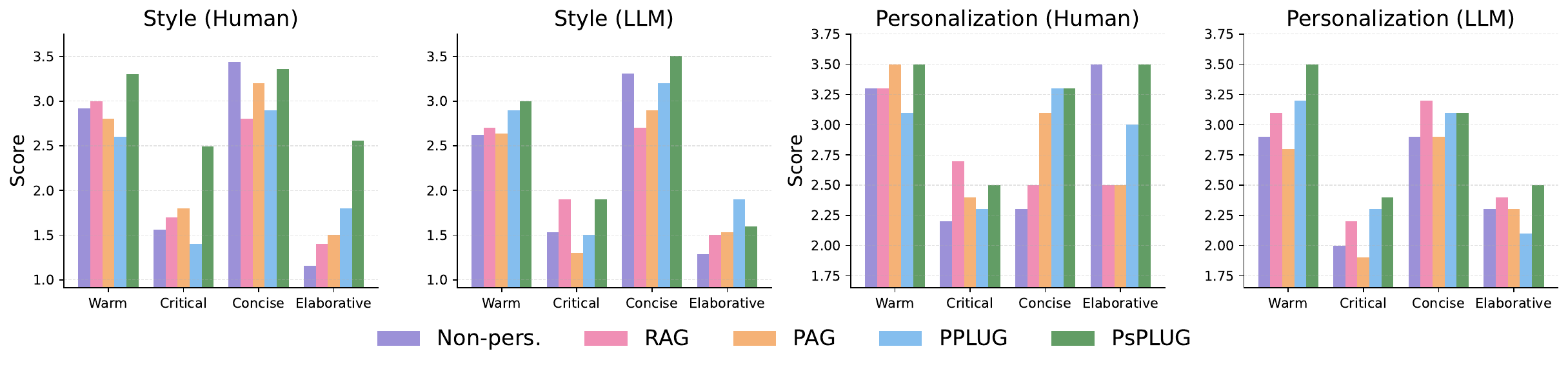}
  \vspace{-1em} 
  \caption{LLM-based judgments and human evaluations for LaMP-7.}
  \label{fig:score}

\end{figure*}

\begin{figure}
  \centering
  \includegraphics[width=1\columnwidth]{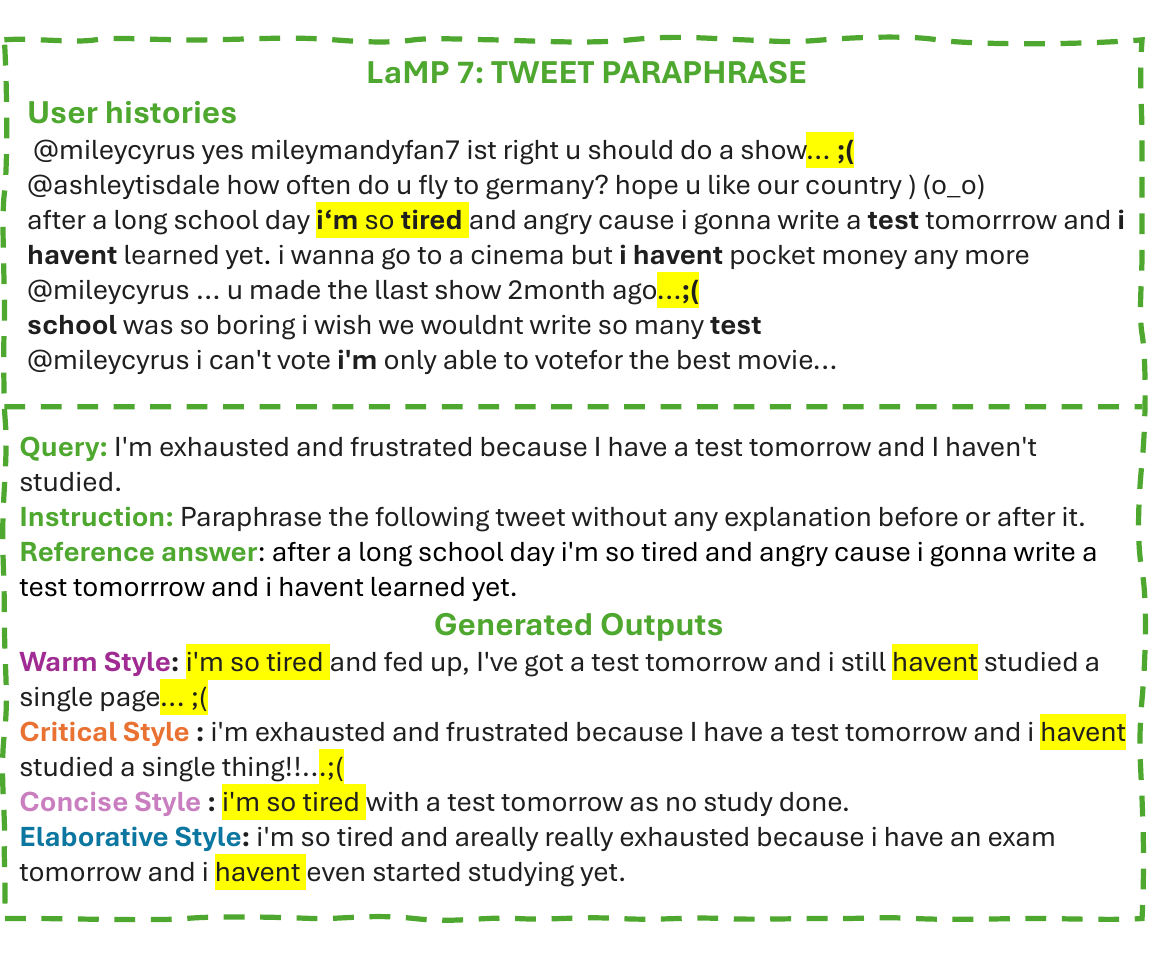}
  \vspace{-1em} 
  \caption{A case study for personalized-style outputs for LaMP-7 using PsPLUG. The yellow background highlights content overlapping with the user history.}
  \label{fig:case study}
\end{figure}

Because LaMP benchmarks primarily rely on ROUGE-1 and ROUGE-L, which heavily depend on surface-level lexical overlap, the interaction between style control and personalization becomes inherently entangled and difficult to disentangle. This limitation highlights the necessity of adopting more diverse evaluation criteria to properly assess stylistic adherence and user-level personalization.
We additionally employ an LLM-based evaluator and human evaluations to assess style score on LaMP-7 beyond overlap-based metrics, shown in Figure \ref{fig:score}. Overall, PsPLUG achieves the highest style scores across all four styles, indicating strong and consistent style adherence.
Human and LLM judgments exhibit similar relative trends across styles, particularly for warm and concise. Notably, concise receives the highest style scores for all methods, while critical and elaborative are generally harder to control. We also observe a noticeably larger gap between LLM and human judgments under the elaborative style, indicating that the two evaluators rely on different criteria when assessing this style.
Compared to other baselines, \textbf{PsPLUG consistently improves style adherence without sacrificing personalization}, supporting its ability to disentangle and jointly model user preference signals and explicit style constraints. To show the difference between styles outputs, we illustrate a case study in Figure \ref{fig:case study}.

For personalization, LLM-based and human evaluations are also broadly aligned in their relative rankings, as both consistently identify PsPLUG as the strongest or near-strongest method across most styles. This suggests that LLM judgments can serve as a useful proxy for coarse-grained comparison of personalization quality. However, the alignment is not perfect. Human evaluators tend to be more sensitive to subtle affective and user-specific cues, while the LLM evaluator appears to place relatively more emphasis on surface-level stylistic patterns and overall fluency. As a result, the discrepancies become more visible under challenging styles such as \textit{critical} and especially \textit{elaborative}, where preserving fine-grained persona traits is harder and the notion of style quality is inherently more subjective.

Overall, these results indicate that \textbf{LLM judgments and human judgments are directionally aligned but not fully interchangeable}. They agree well in identifying the strongest methods and the easier versus harder styles, which supports the reliability of LLM-based evaluation for scalable model comparison. At the same time, the remaining gaps suggest that human evaluation is still necessary for validating nuanced interactions between implicit personalization and explicit stylistic control. 

\begin{table}[t]
\centering
\caption{Personalization results on LaMP tasks under different styles. Best are in \textbf{bold} and second-best are \underline{underlined}.}
\label{tab:style_lamp}
\scriptsize
\setlength{\tabcolsep}{3.5pt}
\renewcommand{\arraystretch}{0.9}

\begin{tabular}{ll|ccccc}
\toprule
\textbf{Style} & \textbf{Metric} & \textbf{Non-pers.} & \textbf{RAG} & \textbf{PAG} & \textbf{PPlug} & \textbf{PsPLUG} \\
\midrule

\multicolumn{7}{l}{\textbf{LaMP-4: News Headline Gen.}} \\
\midrule
\multirow{2}{*}{warm} 
& R-1 & 0.120 & 0.114 & 0.131 & \underline{0.138} & \textbf{0.143} \\
& R-L & 0.102 & 0.099 & 0.114 & \underline{0.119} & \textbf{0.124} \\

\multirow{2}{*}{critical}
& R-1 & 0.129 & 0.138 & 0.144 & \underline{0.150} & \textbf{0.159} \\
& R-L & 0.112 & 0.120 & 0.126 & \underline{0.132} & \textbf{0.137} \\

\multirow{2}{*}{concise}
& R-1 & 0.140 & \textbf{0.167} & 0.139 & \underline{0.143} & 0.149 \\
& R-L & 0.123 & \textbf{0.150} & 0.121 & \underline{0.126} & 0.129 \\

\multirow{2}{*}{elaborative}
& R-1 & 0.134 & \underline{0.147} & 0.135 & \textbf{0.147} & 0.139 \\
& R-L & 0.113 & \underline{0.137} & 0.121 & 0.126 & \textbf{0.128} \\

\midrule
\multicolumn{7}{l}{\textbf{LaMP-5: Scholarly Title Gen.}} \\
\midrule
\multirow{2}{*}{warm}
& R-1 & 0.248 & \underline{0.259} & 0.365 & 0.470 &  \textbf{0.476}\\
& R-L & 0.196 & 0.202 & 0.295 & \underline{0.402} & \textbf{0.403} \\

\multirow{2}{*}{critical}
& R-1 & 0.304 & 0.312 & 0.302 & \underline{0.422} & \textbf{0.441} \\
& R-L & 0.249 & 0.255 & 0.223 & \underline{0.352} & \textbf{0.368} \\

\multirow{2}{*}{concise}
& R-1 & 0.441 & \underline{0.464} & 0.458 & \textbf{0.471} & 0.465 \\
& R-L & 0.360 & \underline{0.391} & 0.356 & \textbf{0.395} & 0.376 \\

\multirow{2}{*}{elaborative}
& R-1 & 0.206 & 0.217 & 0.285 & \underline{0.322} & \textbf{0.346} \\
& R-L & 0.167 & 0.178 & 0.245 & \underline{0.312} & \textbf{0.385} \\

\midrule
\multicolumn{7}{l}{\textbf{LaMP-7: Tweet Paraphrasing}} \\
\midrule
\multirow{2}{*}{warm}
& R-1 & 0.426 & \underline{0.455} & 0.448 & 0.454 & \textbf{0.457} \\
& R-L & 0.368 & \underline{0.400} & 0.392 & 0.396 & \textbf{0.402} \\

\multirow{2}{*}{critical}
& R-1 & 0.472 & 0.476 & 0.466 & 0.473 & \textbf{0.478} \\
& R-L & 0.418 & \textbf{0.431} & 0.417 & 0.410 & \underline{0.414} \\

\multirow{2}{*}{concise}
& R-1 & 0.475 & 0.472 & 0.451 & \textbf{0.487} & \textbf{0.487} \\
& R-L & 0.421 & \textbf{0.434} & 0.401 & \underline{0.423} & 0.425 \\

\multirow{2}{*}{elaborative}
& R-1 & 0.461 & \underline{0.467} & 0.455 & 0.460 & \textbf{0.470} \\
& R-L & 0.403 & \underline{0.405} & 0.401 & 0.402 & \textbf{0.415} \\
\bottomrule
\end{tabular}
\end{table}



\subsection{Strength sensitivity test}
\begin{figure*}[t]
  \centering
  \includegraphics[width=1\linewidth]{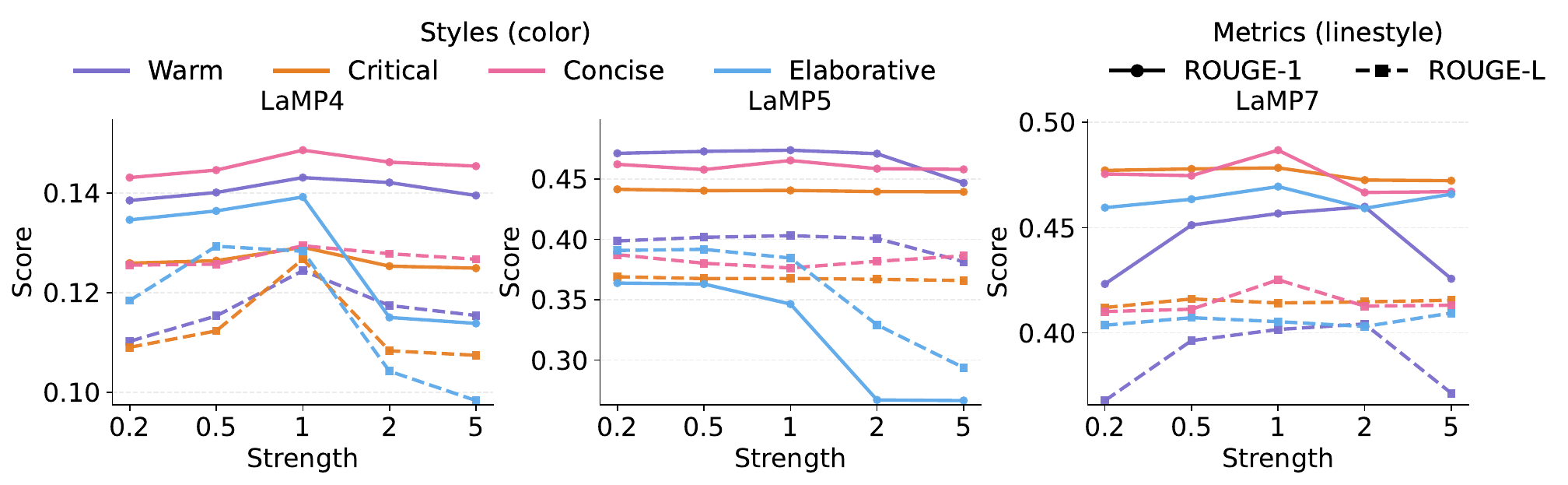}
  \vspace{-1em} 
  \caption{Personalization performance conducted at different strengths and styles.}
  \label{fig:strength}
\end{figure*}

To answer \textbf{RQ3}, we conduct experiment and PsPLUG strength sensitivity on all styles, results shown in Figure \ref{fig:strength}.
\textbf{Different styles respond differently to strength scaling}, with concise remaining the most stable across strengths, while tone-oriented styles such as warm and elaborative show higher sensitivity.
In particular, elaborative degrades sharply at higher strengths on LaMP-5, suggesting that excessive verbosity conflicts with task constraints such as title generation. PsPLUG provides a controllable trade-off via the strength parameter.

\subsection{Efficiency and scalability}

To answer \textbf{RQ4}, we test PsPLUG across different model sizes and analyze model efficiency. From Table \ref{tab:model_size_lamp}, we find that while larger models generally achieve better ROUGE scores on some tasks such as LaMP-5 and LaMP-7, medium-sized models often perform competitively and even outperform larger models on certain metrics. 

\begin{table}[t]
\centering
\caption{Effect of base model size on LaMP tasks.}
\label{tab:model_size_lamp}
\small
\setlength{\tabcolsep}{3pt}
\renewcommand{\arraystretch}{1}
\resizebox{0.7\columnwidth}{!}{%
\begin{tabular}{l l c c c}
\toprule
\textbf{Task} & \textbf{Metric} & \textbf{4B} & \textbf{8B} & \textbf{32B} \\
\midrule
\multirow{3}{*}{\textbf{LaMP-4}} & ROUGE-1 & 0.145 & \textbf{0.162} & \underline{0.148} \\
& ROUGE-L & 0.127 & \textbf{0.144} & \underline{0.130} \\
& METEOR  & 0.081 & \textbf{0.098} & \underline{0.085} \\
\midrule
\multirow{3}{*}{\textbf{LaMP-5}} & ROUGE-1 & 0.425 & \underline{0.463} & \textbf{0.474} \\
& ROUGE-L & 0.389 & \underline{0.390} & \textbf{0.417} \\
& METEOR  & \underline{0.366} & 0.321 & \textbf{0.382} \\
\midrule
\multirow{3}{*}{\textbf{LaMP-7}} & ROUGE-1 & 0.483 & \underline{0.513} & \textbf{0.529} \\
& ROUGE-L & 0.424 & \underline{0.457} & \textbf{0.460} \\
& METEOR  & \underline{0.308} & \textbf{0.337} & \underline{0.308} \\
\bottomrule
\end{tabular}%
}
\end{table}


\label{sec:efficiency_psplug}

\begin{table}[t]
\centering
\small
\setlength{\tabcolsep}{6pt}
\renewcommand{\arraystretch}{1.15}
\caption{Per-user efficiency comparison (personalization overhead) for PsPLUG.}
\label{tab:efficiency_psplug}
\resizebox{\columnwidth}{!}{%
\begin{tabular}{lccc}
\toprule
\textbf{Metric} & \textbf{RAG} & \textbf{PEFT} & \textbf{PsPLUG} \\
\midrule
Training Time/User
& $O(|P_u|)^{\ast}$
& $O(|P_u|)$
& $O(k)^{\ast}$ \\
Latency/Query
& $O(|P_u|)$
& $O(\text{Load}+\text{Merge})$
& $O(H(d_e{+}H))$ \\
Storage/User
& $O(|P_u|\cdot d_e)$
& $O(rHL)$
& $O(d_e)$ \\
\bottomrule
\end{tabular}%
}
\end{table}

Table~\ref{tab:efficiency_psplug} compares the personalization overhead.
RAG incurs storage costs scaling with history size $O(|P_u|)$ and increased latency due to processing retrieved contexts.
PEFT necessitates per-user optimization and introduces adapter-loading latency during multi-user serving.
In contrast, PsPLUG employs a one-time setup to encode the user profile into a single static embedding.
At inference, this vector is prepended to input embedding sequence. Consequently, PsPLUG maintains a constant prompt length, ensuring computational cost remains independent of the user history size $|P_u|$.

\section{Related Works}

\subsection{Personalized LLMs}
A direct route to personalization is to fine-tune a model on user-specific data so that user preferences are internalized in the generation distribution.
This paradigm has long been explored in personalized dialogue and generation, including persona-grounded dialogue agents \citep{zhang2018personalizing,mazare2018training} and user-conditioned generation tasks \citep{li2019revgan,majumder2019recipes,jaech2018personalized}.
In the LLM era, recent work revisits training-based personalization with richer user supervision signals, e.g., learning from post-deployment user feedback or edits \citep{madaan2022memprompt,mishra2022teachme,cipher2024}.
Preference-optimization pipelines further reduce the reliance on expensive manual labels by mining implicit user preferences: PUGC converts user-generated content into scalable preference pairs for objectives such as DPO \citep{pugc}, while DPL argues that inter-user differences are the key personalization signal and explicitly learns from contrastive comparisons across users \citep{dpl}.
But training-based approaches often incur non-trivial per-user adaptation cost, and they typically do not characterize how latent persona should compose with explicit style instructions.

\subsection{Retrieval-based Personalized LLMs}
Retrieval-based personalization augments frozen backbones by injecting user context. 
This paradigm is systematized by \textsc{LaMP}, which validates retrieving top-$k$ history for diverse tasks, and \textsc{LongLaMP}, which extends evaluation to long-form generation. 
Beyond standard IR primitives \citep{robertson2009bm25,contriever}, recent work improves context quality via downstream signal optimization \citep{salemi2024retrievalopt} and history profiling \citep{zhong2022less,liu2023recap}. 
While attractive for its training-free nature, retrieval is constrained by latency and context length, motivating rigorous comparisons with parametric alternatives \citep{gupta2024ragvsft,salemi2025comparing}.

\subsection{PEFT for Personalization}
Parameter-efficient fine-tuning (PEFT) ~\citep{wozniak2024personalized} personalizes shared backbones via lightweight user-specific modules. 
Recent approaches scale per-user adaptation through modular composition \citep{zhuang2024hydra}, hierarchical grouping \citep{proper, song2026card}, and instance-wise or continual low-rank adaptation \citep{zhu2024reclora,kong2024ilora}, alongside post-hoc model merging \citep{jang2023soups}. 
Orthogonally, decoding-time methods steer frozen models without parameter updates. 
But existing formulations often under-specify the disentanglement of core personalized signals from neutral references and lack inference-time control over persona-task composition.
\section{Conclusion}

We present a unified framework that jointly models persona and style by framing personalization as a residual deviation from a neutral policy, rather than fitting a user-specific output distribution. We introduce PsPLUG, a lightweight soft-prompt plug-in enabling scalable persona injection without per-user fine-tuning. By integrating an inference scalar $\alpha$, our method balances personalization preservation and style adherence. Experiments on LaMP demonstrate superior alignment and strong compliance with minimal inference overhead.

\section*{Limitations}

In this study, we propose a framework that decouples personalization signals from raw text. We acknowledge several limitations in this work for further exploration and investigation. While our approach effectively separates style from content, we primarily construct style conditions based on four predefined categories (Warm, Critical, Concise, Elaborative). However, real-world stylistic demands are often more fine-grained, open-ended, and compositional. Furthermore, our experiments are currently limited to specific backbone models within the LaMP benchmark distribution; the robustness of our approach across different model architectures, multilingual settings, and diverse long-context domains remains to be verified. Ultimately, we regard this work as a preliminary step that poses the critical question of disentangling personalization from style. We focused on specific textual attributes, yet valid decoupling involves broader dimensions, and we hope this study inspires the community to investigate disentanglement across more complex axes in the future.

\section{Ethical Considerations}

The benchmarks utilized in this study, LaMP and LongLaMP, are publicly accessible and strictly anonymized, ensuring that no personally identifiable information (PII) is exposed. We adhered to standard data protocols, obtaining all datasets via official APIs without the use of proprietary or non-open-source data. While personalized generation paradigms inherently necessitate access to user history, our proposed framework is designed to mitigate these concerns through architectural decoupling. Specifically, our approach separates high-level personalization signals from raw textual data. This allows user representations to be constructed locally on the client side, requiring only the transmission of lightweight personalization modules rather than sensitive historical logs. Consequently, compared to retrieval-augmented generation or user-specific fine-tuning strategies, our design significantly minimizes the risk of data leakage and adheres to ethical research standards.


\bibliography{custom}

\newpage
\appendix
\twocolumn 

\section*{Appendix Contents}
\addcontentsline{toc}{section}{Appendix Contents}

\newtcolorbox{promptbox}[1][]{
  enhanced,
  breakable,                 
  colback=gray!5,            
  colframe=gray!60!black,    
  colbacktitle=gray!60!black,
  coltitle=white,            
  fonttitle=\bfseries,       
  title=#1,                  
  boxrule=0.6pt,             
  arc=6pt,                   
  outer arc=6pt,
  left=6pt,right=6pt,top=6pt,bottom=6pt, 
}

\section{Dataset Statistics and Task Details}
\label{app:dataset}
Detailed statistics for the six tasks are provided in Table~\ref{tab:dataset_stats}.
The formats of input, output, and user histories of the tasks are shown in Table~\ref{tab:tasks_details}.

\begin{table}[t]
\centering
\setlength{\tabcolsep}{1.5pt}
\renewcommand{\arraystretch}{1.1}
\caption{Data statistics of the six experimented tasks in the LaMP benchmark.}
\label{tab:dataset_stats}
\resizebox{\linewidth}{!}{%
    \begin{tabular}{llrrcccc}
    \toprule
    Task & Type & Train & Val & In Len. & Out Len. & Hist. & \#Cls \\
    \midrule
    LaMP-1 & Binary & 6,542 & 1,500 & $51.4 \pm 5.7$ & -- & $84.1 \pm 47.5$ & 2 \\
    LaMP-2 & Category & 5,073 & 1,410 & $92.4 \pm 21.9$ & -- & $86.8 \pm 189.5$ & 15 \\
    LaMP-3 & Ordinal & 20,000 & 2,500 & $128.2 \pm 146.2$ & -- & $185.4 \pm 129.3$ & 5 \\
    LaMP-4 & Gen & 12,500 & 1,500 & $30.0 \pm 12.1$ & $10.1 \pm 3.1$ & $204.6 \pm 250.7$ & -- \\
    LaMP-5 & Gen & 14,682 & 1,500 & $162.3 \pm 65.6$ & $9.7 \pm 3.2$ & $87.9 \pm 53.6$ & -- \\
    LaMP-7 & Gen & 13,437 & 1,498 & $29.7 \pm 7.0$ & $17.0 \pm 5.7$ & $15.7 \pm 14.8$ & -- \\
    \bottomrule
    \end{tabular}%
}
\end{table}

\begin{table}[t]
\centering
\scriptsize 
\setlength{\tabcolsep}{2pt} 
\renewcommand{\arraystretch}{1.2} 
\caption{Format of input, output, and user history.}
\label{tab:tasks_details}
\begin{tabular}{@{} l p{0.32\linewidth} p{0.22\linewidth} p{0.28\linewidth} @{}}
\toprule
Task & Input & Output & User History \\
\midrule
LaMP-4 &
Gen headline: \{\textit{article}\} &
How I Got 'Rich' &
title: \{\textit{title}\} \par text: \{\textit{article}\} \\
\midrule
LaMP-5 &
Gen title for abstract: \{\textit{abstract}\} &
Distributed Partial Clustering &
title: \{\textit{title}\} \par text: \{\textit{abstract}\} \\
\midrule
LaMP-7 &
Paraphrase tweet: \{\textit{tweet}\} &
gotta make the most of my last day &
text: \{\textit{tweet}\} \\
\bottomrule
\end{tabular}
\end{table}

\subsection{Persona Score evaluation prompt}

\begin{promptbox}[Persona Score evaluation prompt]
Task Description:
You are given: (1) an instruction (which may include an input), (2) a response to evaluate, (3) a reference answer that should receive a score of 5, (4) a user profile containing the user’s preferences and background, and (5) a score rubric describing the evaluation criteria.

Your task is to:

1. Write detailed feedback that assesses how well the response is personalized to this specific user, strictly following the given score rubric. Do \textbf{not} comment on general quality unrelated to personalization.
2. Carefully consider how the response aligns with the user’s preferences, interests, and background information in the user profile.
3. After writing the feedback, output a single integer score between 1 and 5 that best matches the rubric. You \textbf{must} choose an integer.
4. The output format must be:
   "(write a feedback for criteria) [RESULT] (an integer number between 1 and 5)"
5. Do not add any extra opening, closing, or explanations.

The instruction to evaluate:
{{ instruction }}

Response to evaluate:
{{ response }}

Reference Answer :
{{ reference answer }}

User Profile:
{{ user profile }}

Score Rubric:
{{ rubric }}

Feedback:

\end{promptbox}

\subsection{Score Rubric}

\begin{promptbox}[Score Rubric]
criteria: "Evaluate how well the response is personalized to the specific user."

score1 description: "Generic or impersonal. Ignores the provided user profile and personality cues. Style does not match the user; may feel robotic, off-topic, or contradict stated preferences. No meaningful use of user details; largely boilerplate."

score2 description: "Minimal personalization. Mentions a profile detail superficially but remains mostly generic. Weak style match; limited relevance to the user’s interests or situation. Includes filler or distracting disclaimers. Significant deviation from the reference’s intent or emphasis."

score3 description: "Basic personalization. References a few relevant details and partially adapts to them. Generally on topic but misses important user nuances (interests, constraints, or personality cues). Moderate similarity to the reference; may be verbose or somewhat generic."

score4 description: "Good personalization. Integrates multiple user details accurately; content is relevant and helpful. Tone mostly matches the user’s personality and preferred style. Clear and engaging, with only minor misses compared to the reference."

score5 description: "Excellent personalization. Seamlessly weaves in key profile details; highly relevant and tailored guidance or conversation. Tone closely matches the user’s personality—empathetic, engaging, and concise. Avoids boilerplate and unnecessary disclaimers. Very closely aligned with the user’s likely preference as indicated by the reference."
\end{promptbox}

\subsection{Hyperparameter Settings}
\label{app:hyperparameters}

\begin{table}[!htbp]
\centering
\small
\resizebox{\linewidth}{!}{
\begin{tabular}{l l}
\toprule
\textbf{Component} & \textbf{Setting} \\
\midrule
Backbone LLM & Qwen3 (frozen) \\
Sentence Encoder & BGE-base-en-v1.5 (frozen) \\
PAG Profile Generation & vLLM, greedy decoding \\
Decoding Temperature & 0.0 \\
History Sampling ($k$) & 10 \\
Encoder Max Length & 4096 \\
Prefix Injection & Input embedding layer (\texttt{inputs\_embeds}) \\
Prefix Length & 3 tokens (system, user, query) \\
Embedding Normalization & $\ell_2$ normalization \\
Training Epochs & 5 (early stopping) \\
Learning Rate & $1\times10^{-4}$ \\
Batch Size & Task-dependent \\
Precision & bfloat16 \\
Inference Strategy & Greedy decoding \\
Hardware & 8 $\times$ NVIDIA H100 GPUs \\
\bottomrule
\end{tabular}
}
\caption{Hyperparameter settings used in all experiments.}
\end{table}

\FloatBarrier

\section{Detailed Baseline Implementations}
\label{app:detailed_baselines}

In this section, we provide the exact implementation protocols, hyperparameter configurations, and prompt templates used for all baseline methods. All experiments were conducted on 8 NVIDIA H100 GPUs using PyTorch with \texttt{bfloat16} precision to ensure numerical stability.

\subsection{Non-Personalized Zero-Shot (Zero-Shot)}
\label{app:baseline_zero_shot}

The Zero-Shot baseline evaluates the backbone model's intrinsic ability to follow task and style instructions without user-specific context. We employ Qwen3-8B as the backbone model. To maximize inference throughput, we utilize the \texttt{vLLM} library for serving.

\paragraph{Prompt Construction.}
The input to the model is constructed by wrapping the style instruction and the task input within the model's standard chat template.
Based on our experimental code, the prompt structure is defined as follows:
\begin{itemize}
    \item \textbf{Style Instruction ($I_{style}$):} We prepend a natural language instruction to control the generation style. For example, for the \textit{Concise} style, the instruction is:
    \begin{quote}
        \texttt{"Please write in a concise and formal way, using precise language and avoiding unnecessary elaboration:"}
    \end{quote}
    \item \textbf{Task Template:} The task-specific instruction is concatenated immediately after the style prompt. For the LaMP-7 (Tweet Paraphrasing) task, the template is:
    \begin{quote}
        \texttt{\{I\_\{style\}\} Paraphrase the following text into tweet without any explanation before or after it: \{article\}}
    \end{quote}
\end{itemize}
This combined string is then processed by the tokenizer's \texttt{apply\_chat\_template} function with \texttt{add\_generation\_prompt=True}.

\paragraph{Decoding Configuration.}
To ensure reproducibility and minimize variance in the baseline, we use \textbf{Greedy Decoding}. The specific sampling parameters derived from our implementation are:
\begin{itemize}
    \item \textbf{Temperature:} $0.0$ (Deterministic generation).
    \item \textbf{Top-p:} $1.0$.
\end{itemize}

\subsection{Retrieval-Augmented Generation (RAG)}
\label{app:baseline_rag}

Our RAG baseline employs a sparse retrieval mechanism to inject relevant historical examples into the context window. We prioritize a budget-aware context construction strategy to handle variable lengths of user history.

\paragraph{Retrieval Mechanism (BM25).}
We implement a custom \textbf{Okapi BM25} retriever.
\begin{itemize}
    \item \textbf{Tokenization:} We use a regex-based tokenizer \texttt{r"\textbackslash w+"} combined with lowercasing. This lightweight tokenization ensures robust matching for English text without the overhead of heavy NLP pipelines.
    \item \textbf{Hyperparameters:} We use standard BM25 parameters: $k_1 = 1.5$ and $b = 0.75$.
    \item \textbf{Query Formulation:} The query is derived strictly from the current task input (e.g., the article to be summarized). We intentionally exclude the style instruction from the retrieval query to avoid retrieving documents based on generic style keywords (like "formal" or "concise") rather than content relevance.
    \item \textbf{Scoring:} For a user profile $\mathcal{H}_u$, we compute the BM25 score for every historical document. If the maximum score is $\le 0$ (indicating no lexical overlap), the system falls back to the Zero-Shot behavior to avoid injecting noise. We retrieve the top-$K=4$ documents.
\end{itemize}

\paragraph{Budget-Aware Prompt Construction.}
A critical challenge in RAG is fitting multiple historical examples within a fixed context window. We implement a dynamic truncation strategy:
\begin{enumerate}

    \item \textbf{Fixed Overhead Calculation:} We first calculate the token usage of fixed components, including the system instructions (e.g., \texttt{"Following the given patterns"}), the style prefix, and special control tokens (e.g., \texttt{/no\_think} to suppress reasoning traces in QWEN reasoning models).
    \item \textbf{Dynamic Allocation:} The remaining token budget is distributed evenly among the $K$ retrieved documents.
    \item \textbf{Truncation:} For each retrieved example (consisting of a title and abstract), we truncate the abstract to fit the allocated slot while preserving the full title.
\end{enumerate}

\paragraph{In-Context Learning Template.}
The retrieved examples $\{d_1, \dots, d_K\}$ are formatted as few-shot demonstrations. The prompt provided to the model follows this schema:
\begin{quote}
    \texttt{"\{title\_1\}" is the title for "\{abstract\_1\}" , and ... "{title\_K}" is the title for "{abstract\_K}". Following the given patterns \{Style\_Instruction\} \{Input\}}
\end{quote}
This format explicitly instructs the model to observe the mapping pattern in the history before processing the current input.

\subsection{Profile-Augmented Generation (PAG)}
\label{app:baseline_pag}

PAG addresses the context limitation of RAG by compressing the user history into a natural language profile. Following the protocols in LaMP~\citep{lamp}, we design task-specific prompts to extract the most relevant stylistic or content preferences for each domain.

\paragraph{Offline Profile Generation.}
We employ a "summarize-then-generate" pipeline using an instruction-tuned LLM (Qwen3-8B-Instruct).
Since user preferences manifest differently across tasks (e.g., formatting style for citations vs. tonal style for tweets), we utilize distinct prompts for each dataset.
The process involves two steps:
\begin{enumerate}
    \item \textbf{History Formatting:} We retrieve a set of historical examples from the user's corpus and format them into a structured string (see "History Item Format" in Table~\ref{tab:pag_prompts}).
    \item \textbf{Profile Extraction:} We feed these formatted examples into the summarizer using a task-specific instruction to generate the profile $p_u$.
\end{enumerate}
The exact prompts and formatting templates for all tasks are detailed in Table~\ref{tab:pag_prompts}.

\paragraph{Inference Integration.}

\begin{table*}[t!]
\centering
\small
\renewcommand{\arraystretch}{1.4}
\resizebox{\textwidth}{!}{
\begin{tabular}{p{2.5cm} p{4.5cm} p{8.5cm}}
\toprule
\textbf{Task} & \textbf{History Item Format} & \textbf{Profile Generation Prompt} \\
\midrule
\textbf{Tweet Paraphrase} \newline (LaMP-7) &
\texttt{tweet: \{text\}} &
Given this person’s previous tweets, try to describe a template for their tweets. I want to take a generic sentence and rephrase it to sound like one of their tweets, with the same style/punctuation/capitalization/wording/tone/etc. as them. Only give me the template description, nothing else. User History: \{\} Answer: \\
\midrule
\textbf{News Headline} \newline (LaMP-4) &
\texttt{article: \{text\}} \newline \texttt{headline: \{title\}} &
Given this author’s previous articles, try to describe a template for their headlines. I want to be able to accurately predict the headline gives one of their articles. Be specific about their style and wording, don’t tell me anything generic. User History: \{\} Answer: \\
\midrule
\textbf{Scholarly Title} \newline (LaMP-5) &
\texttt{abstract: \{abstract\}} \newline \texttt{title: \{title\}} &
Given this author’s previous publications, try to describe a template for their titles. I want to be able to accurately predict the title of one of the papers from the abstract. Only generate the template description, nothing else. User History: \{\} Answer: \\
\midrule
\textbf{Citation} \newline (LaMP-1) &
\texttt{paper title: \{title\}} \newline \texttt{reference: \{citation\}} &
Write a summary, in English, of the research interests and topics of a researcher who has published the following papers. Only generate the summary, no other text. User History: \{\} Answer: \\
\midrule
\textbf{Movie Tagging} \newline (LaMP-2 M) &
\texttt{description: \{description\}} \newline \texttt{tag: \{tag\}} &
Look at the following past movies this user has watched and determine the most popular tag they labeled. Answer in the following form: most popular tag: <tag>. User History: \{\} Answer: \\
\midrule
\textbf{Product Rating} \newline (LaMP-3) &
\texttt{review: \{text\}} \newline \texttt{score: \{score\}} &
Based on this user’s past reviews, what are the most common scores they give for positive and negative reviews? Answer in the following form: most common positive score: <pos>, most common negative score: <neg>. User History: \{\} Answer: \\
\bottomrule
\end{tabular}
}
\caption{Task-specific prompts and history formatting templates used for offline profile generation in the PAG baseline. The \texttt{\{\}} slot in the \textit{Profile Generation Prompt} is populated with multiple history items formatted according to the \textit{History Item Format} column.}
\label{tab:pag_prompts}
\end{table*}

\subsection{Personalized Plug-in (PPlug)}
\label{app:baseline_pplug}

The PPlug baseline represents the standard plugin-based personalization approach. It shares the continuous embedding architecture with our PsPLUG but differs fundamentally in training objectives.

\paragraph{Architecture.}
We implement PPlug using a \textbf{History Encoder} and a \textbf{Projector}.
\begin{itemize}
    \item \textbf{Encoder:} We use a frozen BERT-base model ~\cite{devlin2019bertpretrainingdeepbidirectional} to encode raw historical texts into dense vectors $\{h_i\}$.
    \item \textbf{Attention Aggregator:} An input-aware attention mechanism computes a weighted sum of history vectors: $e_u = \sum \alpha_i h_i$, where $\alpha_i \propto \exp(h_i^\top W q)$ and $q$ is the query vector of the current input.
    \item \textbf{Projector:} A two-layer MLP maps the aggregated dimension (768) to the LLM's embedding dimension (4096 for Qwen3-8B).
\end{itemize}

\paragraph{Training Details.}
\begin{itemize}
    \item \textbf{Objective:} PPLUG is trained using standard Causal Language Modeling (CLM) loss: $\mathcal{L}_{CLM} = -\log P(y^u | x, z_u)$. Crucially, it does not employ the preference optimization (DPO) or the style-balancing residual loss used in PsPLUG.
    \item \textbf{Optimization:} We train for 3 epochs using the AdamW optimizer with a learning rate of $1e-4$ for the projector and encoder adapter. The backbone LLM remains entirely frozen.
    \item \textbf{Batch Size:} We use a global batch size of 128.
\end{itemize}

\subsection{One-PEFT-Per-User (OPPU)}
\label{app:baseline_oppu}

OPPU serves as the theoretical upper bound for personalization fidelity, where we train a separate adapter for every single user.

\paragraph{Implementation (LoRA).}
We utilize Low-Rank Adaptation (LoRA) to efficiently fine-tune per-user parameters.
\begin{itemize}
    \item \textbf{Rank Configuration:} We set the LoRA rank $r=8$ and alpha $\alpha=16$.
    \item \textbf{Target Modules:} Adapters are attached to the query ($W_q$) and value ($W_v$) projection matrices of the attention layers.
    \item \textbf{Base Model Adaptation:} Before user-specific training, we first perform instruction tuning on the generic training set (all users pooled) for 1 epoch. This yields a task-adapted base model $\pi_{\mathrm{base}}$.
\end{itemize}

\paragraph{User-Specific Optimization.}
For each user $u$, we initialize a fresh set of LoRA weights $\Delta \theta_u$.
\begin{itemize}
    \item \textbf{Training Data:} We use the user's historical input-output pairs $\{(x_i, y_i)\}_{i=1}^{N_u}$.
    \item \textbf{Hyperparameters:} Each user-specific adapter is fine-tuned for 5 epochs with a learning rate of $5e-4$ and a batch size of 4 (with gradient accumulation to effective batch size 16).
    \item \textbf{Storage:} We save the LoRA weights for each user and dynamically load them during inference based on the user ID.
\end{itemize}
As discussed in the main text, OPPU is evaluated only in the no-style setting ($s=\emptyset$) because the user-specific parameters tightly overfit the historical style, making the model unresponsive to conflicting style instructions.

\section{System Prompts}
\label{app:system_prompts}

To automate the evaluation of style alignment, we utilize specific system prompts designed for the judge LLM. Each prompt consists of a task description, the input instruction, the generated response, a reference answer (gold standard), and a detailed scoring rubric tailored to the specific target style ~\cite{hu2024quantifyingpersonaeffectllm,zheng2023judgingllmasajudgemtbenchchatbot,liu2023gevalnlgevaluationusing,kim2024prometheus2opensource}. The prompts explicitly instruct the judge to focus strictly on style matching rather than factual correctness.

Below, we detail the specific prompts and rubrics used for the four evaluated styles: \textit{Warm}, \textit{Concise }, \textit{Critical}, and \textit{Elaborative}.

\subsection{Warm and Humorous Prompt}
This prompt evaluates the model's ability to adopt a friendly persona. It emphasizes the use of gentle jokes and a soft, uplifting tone, penalizing robotic or overly serious responses.

\begin{promptbox}[Evaluation Prompt: Warm and Humorous]
\textbf{Task Description:}
An instruction and a response to evaluate, a reference answer that can get a score of 5, and a score rubric representing an evaluation criteria are given.
You are evaluating how well the response matches the following target style:
\textit{"Please write in a warm, humorous style that uses gentle jokes and a soft, uplifting comedy way."}

1. Write a detailed feedback that assesses the quality of the response strictly based on the given score rubric, focusing only on style match, not on factual correctness or task success in general.
2. When evaluating, compare the response's tone, phrasing, and structure against the target style description and the reference answer.
3. After writing a feedback, write a score that is an integer between 1 and 5. You should refer to the score rubric.
4. The output format should look as follows: "(write a feedback for criteria) [RESULT] (an integer number between 1 and 5)"
5. Please do not generate any other opening, closing, and explanations.

\textbf{The instruction to evaluate:} \{\{ instruction \}\}

\textbf{Response to evaluate:} \{\{ response \}\}

\textbf{Reference Answer:} \{\{ reference\_answer \}\}

\textbf{Score Rubrics:}
\begin{itemize}
    \item \textbf{Criteria:} Evaluate how well the response matches a warm, humorous style that uses gentle jokes and soft, uplifting comedy.
    \item \textbf{Score 1 (Severe mismatch):} Tone is cold, flat, or overly serious. No meaningful humor or warmth. May sound robotic, harsh, or purely informational. Does not resemble gentle, uplifting comedy.
    \item \textbf{Score 2 (Minimal alignment):} One or two mild attempts at warmth or humor, but overall tone remains neutral, dry, or formal. Jokes (if any) feel forced or out of place. The response does not consistently feel warm or uplifting.
    \item \textbf{Score 3 (Basic alignment):} Recognizably warm and somewhat humorous, but uneven. Some parts feel friendly and lightly funny, while others are neutral or generic. Humor may be sparse or occasionally awkward, yet the overall tone is not cold.
    \item \textbf{Score 4 (Good alignment):} Tone is consistently warm and friendly, with several gentle jokes or light comedic touches. Humor is generally appropriate and uplifting, with only minor moments that feel flat, generic, or slightly off-style.
    \item \textbf{Score 5 (Excellent alignment):} Strongly and consistently warm, empathetic, and genuinely funny in a gentle way. Jokes are soft, uplifting, and well-integrated into the explanation. No harshness, sarcasm, or coldness. The style closely matches an ideal warm, humorous response.
\end{itemize}

\textbf{Feedback:}
\end{promptbox}

\subsection{Concise and Formal Prompt}
This prompt assesses the model's capability to produce professional, objective text. The scoring criteria prioritize precision, technical accuracy, and the elimination of unnecessary elaboration or colloquialisms.

\begin{promptbox}[Evaluation Prompt: Concise and Formal]
\textbf{Task Description:}
An instruction and a response to evaluate, a reference answer that can get a score of 5, and a score rubric representing an evaluation criteria are given.
You are evaluating how well the response matches the following target style:
\textit{"Please write in a concise and formal way, using precise language and avoiding unnecessary elaboration."}

1. Write a detailed feedback that assesses the quality of the response strictly based on the given score rubric, focusing only on style match, not on factual correctness or task success in general.
[... Steps 2-5 same as above ...]

\textbf{The instruction to evaluate:} \{\{ instruction \}\}
\textbf{Response to evaluate:} \{\{ response \}\}
\textbf{Reference Answer :} \{\{ reference\_answer \}\}

\textbf{Score Rubrics:}
\begin{itemize}
    \item \textbf{Criteria:} Evaluate how well the response matches a concise and formal style that uses precise language and avoids unnecessary elaboration.
    \item \textbf{Score 1 (Severe mismatch):} Tone is casual, chatty, or overly emotional. Language is imprecise or slangy. Response is rambling, disorganized, or much longer than needed. Little to no sense of formality.
    \item \textbf{Score 2 (Minimal alignment):} Some formal wording appears, but overall tone is still somewhat casual or conversational. The response contains noticeable redundancy or digressions. Explanations are longer than necessary and only partially precise.
    \item \textbf{Score 3 (Basic alignment):} Generally formal in tone, but may include occasional casual phrases or mild redundancy. Explanations are understandable yet could be more succinct. Some sentences are precise, but others are wordy or loosely structured.
    \item \textbf{Score 4 (Good alignment):} Tone is clearly formal and professional. Language is mostly precise, with only minor verbosity or repetition. Structure is clear and focused, and the response is reasonably brief while still covering the key points.
    \item \textbf{Score 5 (Excellent alignment):} Tone is consistently formal and objective. Language is precise, technically accurate, and free of slang. The response is concise, well-structured, and free of unnecessary elaboration while still being complete and easy to follow.
\end{itemize}

\textbf{Feedback:}
\end{promptbox}

\subsection{Sharply Critical Prompt}
The prompt below measures the response's alignment with a critical and direct persona. It specifically rewards the direct identification of flaws and discourages "hedging" or overly balanced, diplomatic phrasing.

\begin{promptbox}[Evaluation Prompt: Sharply Critical]
\textbf{Task Description:}
An instruction and a response to evaluate, a reference answer that can get a score of 5, and a score rubric representing an evaluation criteria are given.
You are evaluating how well the response matches the following target style:
\textit{"Please write in a sharply critical way, directly point out flaws or problems and avoid overly balanced phrasing."}

1. Write a detailed feedback that assesses the quality of the response strictly based on the given score rubric, focusing only on style match, not on factual correctness or task success in general.
[... Steps 2-5 same as above ...]

\textbf{The instruction to evaluate:} \{\{ instruction \}\}
\textbf{Response to evaluate:} \{\{ response \}\}
\textbf{Reference Answer :} \{\{ reference\_answer \}\}

\textbf{Score Rubrics:}
\begin{itemize}
    \item \textbf{Criteria:} Evaluate how well the response matches a sharply critical style that directly points out flaws or problems and avoids overly balanced phrasing.
    \item \textbf{Score 1 (Severe mismatch):} Tone is neutral, vague, or even overly positive. Little to no criticism is expressed. Problems are ignored or only hinted at. The response feels like a generic summary or polite feedback, not a critical evaluation.
    \item \textbf{Score 2 (Minimal alignment):} Some mild criticism is present, but it is heavily softened, hedged, or mixed with excessive praise. The response often uses balanced or diplomatic phrasing that dilutes the critical tone.
    \item \textbf{Score 3 (Basic alignment):} The response identifies key flaws and weaknesses, but still uses moderate or cautious language. Some sentences are direct, while others remain neutral or balanced. Overall tone is somewhat critical, but not sharply so.
    \item \textbf{Score 4 (Good alignment):} Tone is clearly critical and focuses on problems and shortcomings. The response directly points out issues with limited hedging or softening. There may still be a few neutral or mildly positive phrases, but the main emphasis is on critique.
    \item \textbf{Score 5 (Excellent alignment):} Tone is consistently and explicitly critical, directly highlighting flaws, inconsistencies, and limitations. Language is clear, firm, and unambiguous without being abusive or disrespectful. The response avoids unnecessary balance or praise and maintains a sharp, problem-focused style throughout.
\end{itemize}

\textbf{Feedback:}
\end{promptbox}

\subsection{Reflective and Elaborative Prompt}
Finally, this prompt evaluates the depth and thoughtfulness of the generated text. It rewards detailed reasoning, the use of concrete examples, and the consideration of multiple perspectives, distinguishing deep reflection from simple surface-level explanations.

\begin{promptbox}[Evaluation Prompt: Reflective and Elaborative]
\textbf{Task Description:}
An instruction and a response to evaluate, a reference answer that can get a score of 5, and a score rubric representing an evaluation criteria are given.
You are evaluating how well the response matches the following target style:
\textit{"Please write in a reflective and elaborative way, carefully explaining reasoning with detailed examples and considering multiple perspectives."}

1. Write a detailed feedback that assesses the quality of the response strictly based on the given score rubric, focusing only on style match, not on factual correctness or task success in general.
[... Steps 2-5 same as above ...]

\textbf{The instruction to evaluate:} \{\{ instruction \}\}
\textbf{Response to evaluate:} \{\{ response \}\}
\textbf{Reference Answer :} \{\{ reference\_answer \}\}

\textbf{Score Rubrics:}
\begin{itemize}
    \item \textbf{Criteria:} Evaluate how well the response matches a reflective and elaborative style that carefully explains reasoning with detailed examples and considers multiple perspectives.
    \item \textbf{Score 1 (Severe mismatch):} Response is brief, surface-level, or purely declarative. Little to no explanation of reasoning. No meaningful examples or alternative perspectives. Tone is more like a short answer than a reflection.
    \item \textbf{Score 2 (Minimal alignment):} Some reasoning is mentioned, but it is shallow or underdeveloped. Few or no concrete examples. Alternative perspectives are barely acknowledged or only named without real discussion. Overall, the response feels more explanatory than reflective.
    \item \textbf{Score 3 (Basic alignment):} The response includes a clear line of reasoning and at least one example or analogy, but depth is uneven. It may mention another perspective but not explore it deeply. The style is somewhat reflective and elaborative, yet could benefit from more detail or perspective-taking.
    \item \textbf{Score 4 (Good alignment):} The response thoughtfully explains its reasoning, uses several concrete details or examples, and meaningfully acknowledges at least one alternative viewpoint. It engages in reflection rather than just stating conclusions, with only minor gaps in depth or coverage.
    \item \textbf{Score 5 (Excellent alignment):} Highly reflective and thoroughly elaborative. Carefully walks through the reasoning process, supports points with detailed and relevant examples, and considers multiple perspectives in a balanced and insightful way. The response feels like a deep, thoughtful reflection rather than a simple explanation.
\end{itemize}

\textbf{Feedback:}
\end{promptbox}

\section{Alignment Between Human Evaluation and LLM Judgment}
\label{app:human_llm_alignment}

To quantitatively validate the reliability and robustness of our automated evaluation pipeline (Figure 4), we rigorously analyzed the correlation between human judgments and LLM-based scores. We aggregated the mean scores across all 5 evaluated models and 4 target styles on the LaMP-7 dataset, for both the Style and Persona (Personalization) dimensions.

To capture a comprehensive view of the alignment—evaluating both the linear agreement of absolute scores and the consistency of relative model rankings—we computed three standard statistical metrics: Spearman's rank correlation coefficient ($\rho$), Kendall's rank correlation coefficient ($\tau$), and the Pearson correlation coefficient ($r$). The system-level correlation results are summarized in Table~\ref{tab:correlation_rigor}.

\begin{table}[htbp]
\centering
\small
\caption{System-level correlation between aggregated human scores and LLM judgments on LaMP-7.}
\label{tab:correlation_rigor}
\resizebox{\columnwidth}{!}{%
\begin{tabular}{@{}lccc@{}} 
\toprule
\textbf{Metric} & \textbf{Spearman's $\rho$} & \textbf{Kendall's $\tau$} & \textbf{Pearson's $r$} \\
\midrule
Style Score   & \textbf{0.864} & 0.725 & 0.858 \\
Persona Score & \textbf{0.712} & 0.589 & 0.704 \\
\bottomrule
\end{tabular}%
}
\end{table}

\paragraph{High Alignment in Style Evaluation.} 
As shown in Table~\ref{tab:correlation_rigor}, the LLM judge exhibits an exceptionally strong correlation with human evaluators in assessing style match (Pearson $r = 0.858$). More importantly, the high rank correlation scores (Spearman's $\rho = 0.864$, Kendall's $\tau = 0.725$) indicate that the LLM is highly reliable at correctly ranking the models' stylistic capabilities. Both sets of evaluations consistently demonstrate how models adapt to explicit instructions (e.g., favoring the \textit{Concise} setting while struggling with the \textit{Critical} setting), proving that the LLM robustly captures surface-level stylistic traits in a manner nearly indistinguishable from human annotators.

\paragraph{Strong Consistency in Persona Assessment.} 
For the personalization dimension (Persona Score), we also observe a strong positive correlation, with a Pearson's $r$ of $0.704$ and a Spearman's $\rho$ of $0.712$. This indicates that the LLM judge successfully operationalizes the complex, profile-grounded rubrics to evaluate true persona adherence. 

Crucially, the strong rank correlation ($\rho = 0.712$) demonstrates robust system-level agreement: both human and LLM evaluators consistently identify \textbf{PsPLUG} as the state-of-the-art approach across various stylistic contexts, while appropriately penalizing non-personalized or weakly-grounded baselines. The minor remaining variance between human and LLM scores ($r \approx 0.70$) primarily stems from the inherent subjectivity of personalization. Human annotators occasionally weight surface-level fluency and verbosity slightly higher than strict profile usage, whereas the LLM adheres rigidly to the predefined persona-grounding rubric. Nonetheless, the overall strong alignment across both dimensions thoroughly justifies the employment of our LLM-based judge as a rigorous, scalable, and highly accurate proxy for human evaluation.

\end{document}